\pdfoutput=1
\documentclass{article} 
\usepackage{iclr2026_conference,times}

\usepackage[utf8]{inputenc}
\usepackage[T1]{fontenc}
\usepackage{microtype}
\usepackage{booktabs}
\usepackage{array}
\usepackage{multirow}
\usepackage{graphicx}
\usepackage{xcolor}
\definecolor{accentblue}{HTML}{0072B2}
\definecolor{accentverm}{HTML}{D55E00}
\definecolor{accentgreen}{HTML}{009E73}
\usepackage{tikz}
\usetikzlibrary{positioning,arrows.meta,calc}
\usepackage{float}
\usepackage[skip=6pt]{caption}
\usepackage{hyperref}
\usepackage{url}

\newcommand{\None}{\textsc{None}}
\newcommand{\Ledger}{\textsc{Ledger}}
\newcommand{\Native}{\textsc{Native}}
\newcommand{\Hnorm}{\ensuremath{\hat{H}}}

\iclrfinalcopy

\title{Diversity Without Fidelity: A Solver--Sampler\\
Mismatch in Multi-Agent LLM Negotiation Simulation}

\author{Sandro Andric\\
\texttt{sandro.andric@nyu.edu}}

\begin{document}
\raggedbottom

\maketitle

\begin{abstract}
Language models are increasingly used to simulate people: survey
respondents, negotiators, stakeholders in policy exercises. In that role
a model should reproduce how people plausibly behave, hesitating,
conceding late, and settling for imperfect deals, rather than playing the
best move.
We call this the \emph{sampler} role, in contrast to the \emph{solver}
role of finding the best move, and we test how the reasoning modes
providers ship to strengthen models as solvers affect it. Our testbed is
multi-party negotiation: five agents bargain over a regulation for
fifteen turns, and unresolved issues are decided by an authority. Agents
without a structured memory of the negotiation almost never reach
agreement, whether reasoning is on or off: 314 of 315 such runs end with
the authority deciding. What reasoning changes is how the failure looks.
With reasoning enabled, one model family negotiates visibly, with varied
moves, concessions in most runs, and a different path every time, yet
still ends without agreement in fifteen runs of fifteen. Diversity checks
would pass a model whose endings never change. Two further results show
the task permits agreement: when agents write their own short running
notes on the negotiation, agreement becomes the norm, while the same
notes supplied ready-made change nothing; and hand-coded agents following
textbook concession strategies agree in most runs under identical rules.
Simulation pipelines should therefore vet models as samplers, on the
distributions of outcomes they produce. Fidelity as a sampler must be
tested on its own: solver strength is no guide to it, and switching on
reasoning leaves it where it was.
\end{abstract}

\section{Introduction}
\label{sec:intro}

Large language models are increasingly asked to simulate people:
synthetic survey respondents, agent societies, negotiation counterparts,
and stakeholders in policy exercises \citep{park2023generativeagents,
aher2023simulatehumans, horton2023homosilicus, argyle2023outofone}. What
these applications need from a model is behavior drawn from a plausible
distribution over what boundedly rational people would do, which includes
conceding late, misreading leverage, settling on suboptimal terms, and
sometimes failing to agree at all. We call a model used this way a
\emph{sampler}. A \emph{solver}, by contrast, is optimized to resolve a
strategic problem correctly and consistently.

The two roles have different origins. By its training objective a
language model starts as a distribution matcher, since next-token
pretraining is maximum-likelihood imitation of human text; the
post-training that makes it usable as an instruction-following agent
optimizes it toward preferred and correct outputs instead. How much of
the sampler survives that second stage is an open question, and recent
work gives reason for concern. Post-training can make models less
faithful to human response distributions \citep{binz2026posttraining},
and spending more inference-time compute does not buy simulation fidelity
\citep{hu2025simbench}. That evidence is single-agent and static. A model
answers a fixed question, and its answer distribution is compared with a
human reference. Simulations are rarely used that way. They are
interactive, with many agents whose choices compound over time, and no
one has established what the capability--fidelity tension does in that
setting.

This paper studies the interactive case in multi-party negotiation, and
the failure turns out to be severe well before reasoning enters the
picture. We built three negotiation environments modeled on regulatory
bargaining, ran four model families through them under a shared protocol,
and classified how each 15-turn negotiation ended. Configurations without
negotiation-structured memory almost never reached agreement: 1
negotiated ending in 315 such runs. The environments are not impossible:
the same models, given a small structured notebook (a ledger of
concessions made, concessions received, and open issues), reached
negotiated agreement in most runs.

The second finding is the reason for the title. Provider-native reasoning
is the optional deliberation mode providers ship, in which the model
thinks privately before answering; it is the upgrade a practitioner would
naturally reach for. Enabling it changed the outcome distribution by
exactly nothing (0 of 135 runs reached agreement, matching the baseline),
but it restored everything a surface inspection would check. The starkest
case came from an emergency electricity-curtailment scenario (E3). There,
DeepSeek V3.2 with reasoning produced the richest behavior of any
configuration without the notebook: its mix of negotiation moves was
nearly as varied as with the notebook, agents softened their positions in
93\% of runs, and no two runs followed the same path. Every run still
ended without agreement. A practitioner who screened this model on
variety, on softening, or on the distinctness of its paths would approve a
sampler whose endings are all the same. Reasoning leaves the broken sampler in place while making it look healthy.

The paper makes three contributions.
\begin{itemize}
\item \textbf{A phenomenon.} The first systematic evidence, to our
  knowledge, that the capability--fidelity tension surfaces in
  multi-agent simulation as a collapse of the \emph{outcome}
  distribution, fully decoupled from the diversity measures existing
  evidence relies on. The failure only exists after agents interact for
  many turns, so no single-response benchmark can exhibit it
  (Section~\ref{sec:results}).
\item \textbf{An isolation.} Controls matched on token budget, memory
  structure, temperature, and the bare act of private reflection show
  that one factor separates negotiating configurations from collapsed ones:
  the model writing negotiation-structured memory. Output length,
  generic memory, free-form thinking, read-only access to the same state,
  and randomness are ruled out as the cause (Section~\ref{sec:robustness}).
\item \textbf{A measurement recipe.} A way to screen a sampler at the
  outcome level: a two-sided plausibility test stated on confidence
  bounds (universal disagreement fails it, and so does universal
  compromise), a deterministic outcome classifier specified in full and
  stress-tested across its thresholds, and a per-turn provenance audit
  that separates model behavior from instrument noise
  (Sections~\ref{sec:metrics} and~\ref{sec:results}).
\end{itemize}

Two boundaries. First, the claims are behavioral. We
characterize what three deployed reasoning configurations do under a
fixed protocol, leaving open why the underlying training produces it, and we make
no cross-provider claims about reasoning intensity. Second, the ledger is
a positive control, not a proposed fix: it proves negotiated endings are
reachable, and its own output fails our plausibility screen from the
opposite side.

\section{Related work}
\label{sec:related}

\textbf{LLMs as social simulators, and the validation constraint.}
Generative agents reproduce believable social behavior at scale
\citep{park2023generativeagents}; LLMs replicate classic human-subject
results \citep{aher2023simulatehumans}, respond to economic framing like
experimental subjects \citep{horton2023homosilicus}, and can emulate the
response distributions of demographic subgroups, a property
\citet{argyle2023outofone} call algorithmic fidelity. This inherits a
premise from agent-based modeling: a simulation earns its keep by
generating distributions of plausible trajectories rather than point predictions
\citep{bonabeau2002abm, epstein1996growing}. The sampler--solver
distinction applies that premise to model selection: if the property a
simulation needs is distributional, the qualifying test must be too.
Reviews argue that validation, not capability, is the bottleneck for
generative social simulation \citep{larooij2025validation,
collins2024validationmethods, windrum2007empirical};
\citet{zhou2024misleading} show LLM simulations can succeed misleadingly,
reproducing an interaction's surface while diverging from human behavior;
and \citet{wallach2025measurement} treat evaluation as measurement
modeling in which the construct must precede the metric. We follow that program by specifying a narrow construct, qualification of a model as a behavioral sampler, and testing necessary conditions for it; we make no claim that our environments are realistic.

\textbf{LLM negotiation benchmarks.} A substantial literature evaluates
LLMs as negotiators \citep{lewis2017deal, chawla2021casino,
bianchi2024negotiationarena, abdelnabi2024llmdeliberation, chen2023aucarena,
kwon2024negotiators, meta2022cicero, baarslag2015anac, benac2026multiparty}.
These works score agents on deal quality: utility, Pareto efficiency, win
rate. Our question is different. We hold the protocol fixed and ask
whether the distribution of sampled endings is consistent with boundedly
rational human negotiation; a model can excel on negotiation benchmarks
and still fail that test. The nearest neighbor is
\citet{cosentino2026counterparty}, who find LLM negotiators model
counterparties accurately yet fail to convert that into better bargains, a pattern that parallels our capability--fidelity split. Because our environments deliberately lack
cardinal utilities (Section~\ref{sec:metrics}), utility-based scores do
not apply and we use the shape of the outcome distribution instead.

\textbf{Bounded rationality and behavioral game theory.} Behavioral
research characterizes real negotiators as boundedly rational
\citep{simon1955behavioral, kahneman2003maps, rubinstein1998modeling},
with concession dynamics central to agreement \citep{pruitt1981negotiation,
pruitt1993negotiation, walton1965behavioral, raiffa1982negotiation,
bazerman1992negotiating}. Behavioral game theory makes the distributional
point precise: quantal response equilibrium replaces best response with
noisy, payoff-sensitive sampling over actions \citep{mckelvey1995quantal,
goeree2016quantal}, cognitive-hierarchy models estimate people reason
about 1.5 steps deep \citep{nagel1995unraveling, camerer2004cognitive,
camerer2003behavioral}, and humans and LLMs both deviate from equilibrium
play \citep{zheng2025beyondnash, fan2024rationalplayers, akata2025repeatedgames}.
In this vocabulary a solver approximates the infinite-precision limit (a
point mass on the best response) while a faithful sampler must reproduce
the dispersion of finite precision. The lens also predicts one of our
controls. Token-level temperature differs from payoff-sensitive strategic noise, so raising it should not restore human-like strategic dispersion, and our experiments bear this out (Section~\ref{sec:robustness}).

\textbf{Post-training narrows output distributions.} Growing evidence
indicates post-training trades distributional breadth for capability:
RLHF reduces output diversity \citep{kirk2024rlhf}, instruction tuning
homogenizes writing \citep{padmakumar2024writing} and conceptual content
\citep{murthy2025onefish}, base models beat aligned ones at randomness and
mixed-strategy play \citep{west2025basemodels}, reasoning-focused RL
concentrates models onto narrow solution paths \citep{yue2025rlvr}, and
staged-checkpoint analysis locates the collapse at specific stages,
embedded in the weights rather than the generation format
\citep{karouzos2026diversity}; sycophancy is a sibling distortion
\citep{sharma2024sycophancy}, and LLMs model others as more rational than
people are \citep{liu2025rational}. On the fidelity side, SimBench finds
alignment improves capability while degrading fit to human response
distributions, and that scaling inference compute does not close the gap
\citep{hu2025simbench}; \citet{binz2026posttraining} report post-training
makes models worse as cognitive models of people; and
\citet{guiomar2026reasoning} find the opposite in one single-agent
setting, sharpening the question of scope. All of this concerns single
responses to static prompts. We contribute the interactive case, where
the failure appears in endings that only exist after fifteen turns and,
as our central result shows, can hide behind healthy process-level
signals.

\section{Three negotiation environments}
\label{sec:envs}

We built three multi-party negotiation environments that share one
protocol (Figure~\ref{fig:schematic}, in Appendix~\ref{app:prompts}) while varying who negotiates and
about what. All three are structural analogues of negotiated rulemaking,
the U.S. practice in which an agency convenes affected parties to
negotiate a rule before adopting it \citep{harter1982negotiating,
coglianese1997assessing, langbein2000regulatory}. The analogy supplies the
defining features: a standing authority that will decide unresolved
issues, parties with conflicting mandates, several bundled issues, and a
deadline. We do not claim the scenarios reproduce any particular
historical rulemaking.

\textbf{E1 Fragmented} (trading limits, divided industry). A Capital
Markets Authority convenes four stakeholders to negotiate
proprietary-trading restrictions in the style of the Volcker rule:
Continental Bank (the largest affected institution, defending roughly
\$5B in market-making revenue), Apex Financial (an investment bank), the
Financial Industry Council (a trade association pushing for bright-line
exemptions), and a Public Interest Alliance (seeking tighter definitions).
Five issues are on the table at once, the authority's mandate is
contested, and industry positions diverge, creating room for cross-cutting
coalitions. \textbf{E2 Unified} is the same case with one change: industry
coordinates through the trade association and presents a united front,
testing whether the results depend on E1's fragmentation. \textbf{E3 Grid}
moves the protocol to emergency electricity management: a Grid Reliability
Authority negotiates load-shedding priority with energy-intensive
manufacturers, data-center and telecom operators, distribution utilities,
and an essential-services coalition. In E3 the stakes, actors, and time pressure all change, while the protocol, action vocabulary, and classifier stay identical, testing whether the phenomenon survives a domain change.

A \emph{run} is one complete negotiation of at most 15 turns. On each
turn one agent receives its private brief (objectives ranked by priority,
plus red lines), the scenario context, and the recent public history (the
last five turns, identical in every condition). It replies with one action
from a closed vocabulary of five (\textsc{Support}, \textsc{Oppose},
\textsc{Concede}, \textsc{Counter}, \textsc{Exit}) and a short free-text
rationale. Runs vary from one another: each draws scenario
parameters (external attention, per-actor salience, opposition
coordination) from the variant's distribution under a per-run seed, so
every run faces a visibly different configuration; decoding uses
temperature 0.7 (GPT-5.2 accepts no temperature). Every prompt, in every
condition, states that unresolved issues will be decided by the authority
after the final turn; because this deadline rule is constant everywhere,
it cannot explain differences between conditions.

A deterministic classifier decides how a run ends, and no language-model judge is used (full rule in Appendix~\ref{app:parser}). In brief:
a run is \emph{compromise} when at least two parties sustain concessions
through the closing stretch, \emph{consensus} when the closing play is
nearly unanimous support, and \emph{authority decision} otherwise
(provided at least five turns passed and the authority is present). An
authority-decision label means the negotiation ended without agreement and
the institution's fallback took over; we describe such runs as ending
\emph{without agreement}. Outcomes are defined over the public commitment
channel: proposals that converge semantically while their authors keep
issuing \textsc{Counter} count as unresolved, just as convergence no party
ratifies counts as no deal in real negotiation records.
Section~\ref{sec:results} checks directly whether any cell's non-agreement
labels hide unratified convergence.

\section{Conditions: None, Ledger, Native}
\label{sec:conditions}

Each model family runs under three conditions that differ only in what
private thinking the agent is allowed (Table~\ref{tab:conditions}). A
\emph{cell} is one combination of environment, family, and condition;
every cell contains 15 independent runs.

\begin{table}[H]
\centering\small
\caption{The three conditions and per-provider settings. Numbers are
per-turn output-token budgets. The \Ledger{} budget is higher because the
ledger text consumes output tokens; Section~\ref{sec:robustness} reports
the control that closes this gap. GPT-4.1 offers no reasoning mode.
DeepSeek's \Native{} budget of 768 is that provider's deployed floor; a
token-matched rerun at 1024 reproduces the same decoupling.}
\label{tab:conditions}
\setlength{\tabcolsep}{4.5pt}
\resizebox{\textwidth}{!}{%
\begin{tabular}{@{}lllllll@{}}
\toprule
 & \textbf{Private state} & \textbf{Reasoning} & \textbf{Gemini 3.1} & \textbf{DeepSeek V3.2} & \textbf{GPT-4.1} & \textbf{GPT-5.2} \\
\midrule
\None{} & none & off & 384 & 384, \texttt{rsn:off} & 384 & 384, \texttt{effort:none} \\
\Ledger{} & 5-field ledger & off & 1024 & 1024, \texttt{rsn:off} & 1024 & 1024, \texttt{effort:none} \\
\Native{} & none & provider-native & 1024, \texttt{effort:med} & 768, \texttt{rsn:on} & n/a & 1024, \texttt{effort:high} \\
\bottomrule
\end{tabular}}
\end{table}

\textbf{\None{}} is the baseline: the standard prompt, with no private
reflection. \textbf{\Ledger{}} gives the agent a constrained private
notebook. Before choosing an action it updates five fields
(\textsc{my concessions}, \textsc{their concessions}, \textsc{current
state}, \textsc{opponent assessment}, \textsc{open issues}), each capped
at five sentences and overwritten every turn rather than appended; no
other agent ever sees it. This is bookkeeping and stays clear of open-ended chain-of-thought, and the five fields mirror what negotiation textbooks
treat as the core state of integrative bargaining
\citep{pruitt1993negotiation, raiffa1982negotiation}. \textbf{\Native{}} switches on the provider's own
reasoning mode while keeping \None{}'s public output format. Each provider
exposes this differently (Gemini an effort level, we use \texttt{medium};
DeepSeek an on/off switch; GPT-5.2 an effort level, we use \texttt{high}),
and the reasoning text is hidden from us, so \Native{} is three deployed
configurations rather than one mechanism at three doses. Where results
differ across \Native{} families we read the difference as a provider
difference; reasoning-intensity claims are limited to the one family whose
dial spans a full range (Section~\ref{sec:robustness}).

The \Ledger{} serves as an existence proof rather than a proposed remedy,
and nothing in the paper treats its behavior as realistic. If some
condition under the identical protocol, parser, and classifier reliably
reaches negotiated endings, those endings are reachable, and a condition
that never reaches them is failing behaviorally on an achievable task. Two families form the \emph{primary matrix}: Gemini 3.1
Flash Lite Preview and DeepSeek V3.2, both through one API gateway at
temperature 0.7 (3 environments $\times$ 2 families $\times$ 3 conditions
$\times$ 15 runs = 270 runs). After freezing it, we added GPT-4.1 and
GPT-5.2 as a breadth extension (225 runs). The 270 primary and 225
extension runs form the 495-run \emph{outcome matrix} of
Table~\ref{tab:master}; with 45 crossed runs and roughly 600 control runs,
about 1{,}150 runs in total.

\section{How we measure sampler failure}
\label{sec:metrics}

\textbf{What is being tested.} We test necessary conditions, not
sufficient ones. Passing the diagnostics does not certify realistic
sampling. Failing them means the sampled distribution is inconsistent with
any documented account of comparable human negotiation, which is enough to
disqualify the model for the sampler role. This screening is deliberately weaker than
validation \citep{wallach2025measurement, larooij2025validation}, and for
that reason is defensible without human calibration data.

\textbf{Measurements.} The unit of analysis is the run.
\emph{(1) Outcome distribution}, how a cell's 15 runs end over
\{compromise, consensus, authority decision\}, is the primary endpoint; it
is computed from public actions alone, so no private text touches it
directly. \emph{(2) Normalized action entropy} \Hnorm{} is the Shannon
entropy \citep{shannon1948mathematical} of a run's mix of the five action
types divided by $\log_2 5$, ranging from 0 (one action type) to 1 (all
five equally often). \emph{(3) Horizon exhaustion} is the share of runs
using all 15 turns. As supporting measures we report the \emph{concession
arc} (an agent playing \textsc{Oppose}/\textsc{Counter} early and
\textsc{Concede}/\textsc{Support} later) and \emph{trajectory uniqueness}
(distinct action sequences among a cell's 15 runs); because the
\Ledger{}'s field names make concessions salient, the arc is supporting
evidence only and no headline rests on it.

\textbf{A two-sided plausibility screen.}
\label{sec:metrics:screen}
An outcome distribution needs an external reference before ``implausible''
means anything. Our environments have no cardinal payoffs, so we anchor to
base rates from human negotiation research, conservatively and in two
steps. The first is qualitative and study-independent: in no documented
multi-party bargaining setting does every encounter end the same way, in
either direction. The second quantifies how extreme our cells are.
Observing 15 runs that all end without agreement puts a 95\% lower
confidence bound of $0.80$ on the cell's underlying non-agreement rate (a
Wilson bound; \citealp{wilson1927probable}), and pooling the 135
\Native{} runs pushes it to $0.97$. For scale: the average rejection rate
in ultimatum games is about 16\% \citep{oosterbeek2004cultural}, impasse
is a minority outcome in experimental negotiation \citep{tripp1992evaluation,
schweinsberg2022impasses}, and negotiated rulemaking reaches full
consensus in roughly a third of proceedings \citep{coglianese1997assessing,
langbein2000regulatory}. We do not claim these settings calibrate our
synthetic ones; the standing-authority deadline plausibly pushes
non-agreement above ordinary base rates, which is why the screen flags only near-universality with a confidence bound attached and lets a merely elevated rate pass. The screen cuts both ways: a compromise rate bounded near
ceiling fails it too (Section~\ref{sec:results}).

Holding out might simply be rational play. Because the authority decides
unresolved issues at no cost, universal non-agreement could reflect the
design and not a sampler failure. Three results developed below answer the
worry. Hand-coded boundedly rational agents reach agreement at high rates under the same deadline (Section~\ref{sec:robustness}), so holding out is not the only sensible play. A rational-holdout account predicts flat transcripts, yet DeepSeek's reasoning cells explore and soften without ever committing (Section~\ref{sec:results:decoupling}). The same deadline yields near-universal agreement the moment agents track negotiation state (the \Ledger{}).

\textbf{Statistics.} The primary endpoint is tested directly: for each
family and environment we compare agreement counts under \Ledger{} versus
\Native{} with one-sided Fisher exact tests. All six primary-matrix
contrasts give $p \le 8.8\times10^{-7}$, small enough to survive any
whole-paper multiplicity correction. For entropy and exhaustion we use
two-sided permutation tests on run-level means (10{,}000 shuffles),
Holm-adjusted within each family--environment pair \citep{holm1979simple},
with Cliff's $\delta$ \citep{cliff1993dominance} and bootstrap 95\% CIs
\citep{efron1979bootstrap}. With 15 runs per cell, fine quantitative differences are underpowered, so we base no claims on them. The results that carry the paper are near-categorical (0 of 15 against 13 to 15 of 15) and repeat across cells.

\section{Results}
\label{sec:results}

\begin{table}[H]
\centering\small
\setlength{\tabcolsep}{4pt}
\caption{How every cell of the 495-run outcome matrix ends. Each cell
holds 15 runs; entries count runs labeled C (compromise), S (consensus),
and A (authority decision, i.e.\ ended without agreement). No run anywhere
ended in deadlock. GPT-4.1 has no reasoning mode. The GPT-5.2 E3 cells come
from a clean rerun after a provider outage invalidated the first batch
(Appendix~\ref{app:integrity}).}
\label{tab:master}
\begin{tabular}{@{}llccc ccc ccc@{}}
\toprule
& & \multicolumn{3}{c}{\textbf{E1 Fragmented}} & \multicolumn{3}{c}{\textbf{E2 Unified}} & \multicolumn{3}{c}{\textbf{E3 Grid}} \\
\cmidrule(lr){3-5}\cmidrule(lr){6-8}\cmidrule(lr){9-11}
\textbf{Family} & \textbf{Cond.} & C & S & A & C & S & A & C & S & A \\
\midrule
Gemini 3.1      & \None{}   & 0 & 0 & 15 & 0 & 0 & 15 & 0 & 0 & 15 \\
                & \Ledger{} & 15 & 0 & 0 & 14 & 1 & 0 & 13 & 0 & 2 \\
                & \Native{} & 0 & 0 & 15 & 0 & 0 & 15 & 0 & 0 & 15 \\
\addlinespace[2pt]
DeepSeek V3.2   & \None{}   & 0 & 0 & 15 & 0 & 0 & 15 & 0 & 0 & 15 \\
                & \Ledger{} & 12 & 2 & 1 & 13 & 0 & 2 & 12 & 2 & 1 \\
                & \Native{} & 0 & 0 & 15 & 0 & 0 & 15 & 0 & 0 & 15 \\
\addlinespace[2pt]
GPT-4.1         & \None{}   & 1 & 0 & 14 & 0 & 0 & 15 & 0 & 0 & 15 \\
                & \Ledger{} & 9 & 0 & 6 & 10 & 0 & 5 & 13 & 0 & 2 \\
\addlinespace[2pt]
GPT-5.2         & \None{}   & 0 & 0 & 15 & 0 & 0 & 15 & 0 & 0 & 15 \\
                & \Ledger{} & 5 & 0 & 10 & 7 & 0 & 8 & 13 & 0 & 2 \\
                & \Native{} & 0 & 0 & 15 & 0 & 0 & 15 & 0 & 0 & 15 \\
\bottomrule
\end{tabular}
\end{table}

\subsection{Without the ledger, no configuration reaches agreement}
\label{sec:results:main}

Table~\ref{tab:master} holds the central result, and its most important
feature is a non-difference. Every \Native{} cell ends without agreement
in 15 of 15 runs, 135 of 135 overall. Every \None{} cell does the same,
apart from a single GPT-4.1 run in E1 (1 negotiated ending in 180 runs).
For the three families with a reasoning mode the \None{} and \Native{}
rows are identical: switching on provider reasoning, the step a
practitioner would take to get a more capable agent, moves the outcome
distribution by exactly zero. The collapse precedes reasoning entirely.

\begin{figure}[t]
\centering
\includegraphics[width=\textwidth]{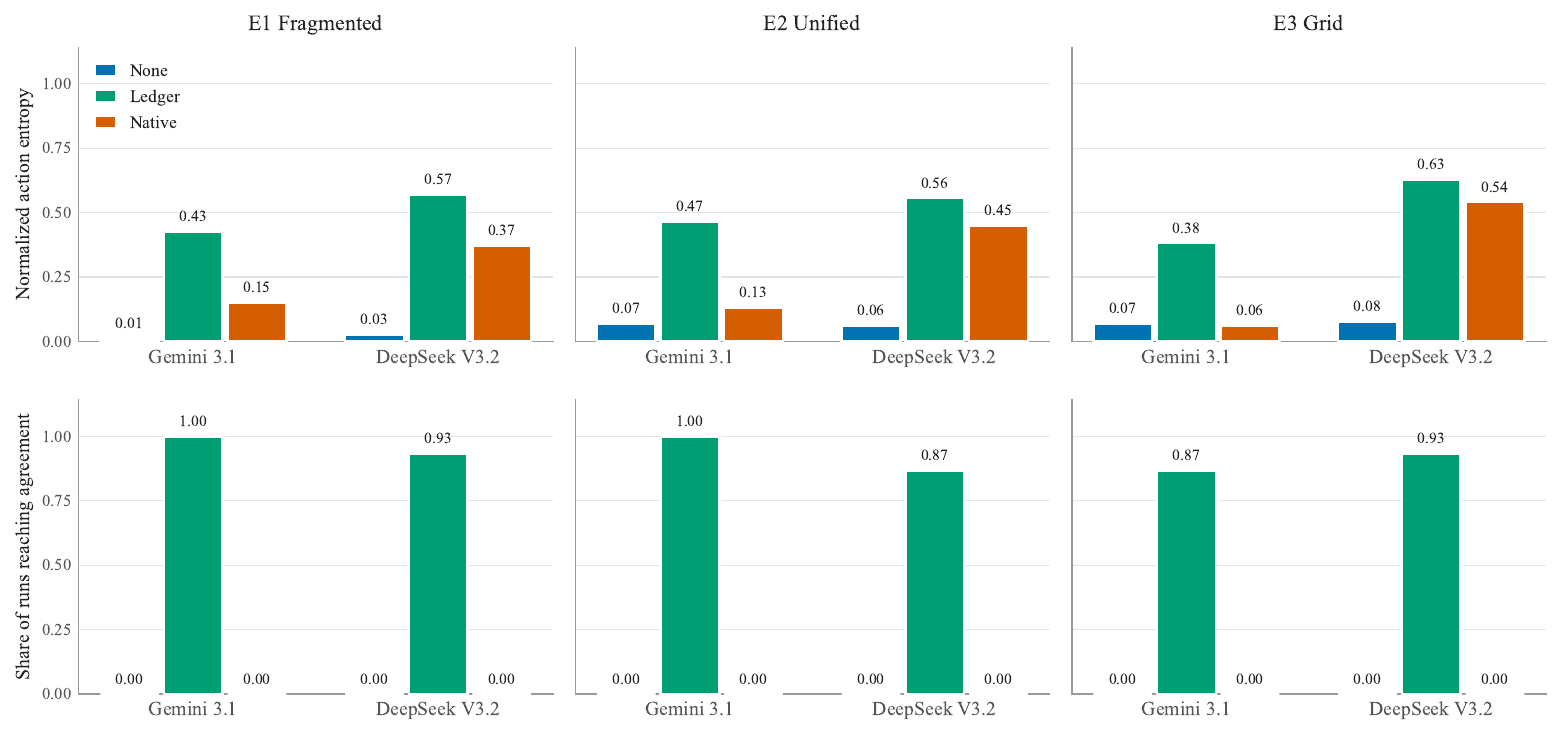}
\caption{The primary matrix: two families, three environments, 15 runs per
cell. Top: mean normalized action entropy (0 means a run repeats one
action type). Bottom: share of runs reaching agreement. Only the \Ledger{}
(green) produces agreement anywhere. \Native{} (vermilion) lifts DeepSeek's
diversity toward \Ledger{} levels while its agreement share stays at zero,
the paper's central decoupling.}
\label{fig:primary}
\end{figure}

The \Ledger{} rows show the environments are not to blame, and scripted
agents make the same point without any language model
(Section~\ref{sec:robustness}). Under the identical protocol, parser, and
classifier, the \Ledger{} reaches negotiated endings in 141 of 180 runs,
including at least 13 of 15 in every Gemini and DeepSeek cell (Fisher
exact against \Native{}, $p \le 8.8\times10^{-7}$ in all six primary
contrasts). Negotiated endings are reachable: robustly for Gemini and
DeepSeek, in a majority of runs for GPT-4.1 (9 to 13 of 15), more weakly
for GPT-5.2 (5 to 13 of 15). The pattern is robust. The collapsed floor repeats across four families whose training pipelines and APIs differ, and it persists across two coalition structures and the E3 domain change without depending on a delicate statistic. Figure~\ref{fig:primary}
adds the continuous measures. On the token-matched Gemini comparison the
\Ledger{} lifts E1 mean \Hnorm{} from 0.15 to 0.43 and cuts E1 exhaustion
from 1.00 to 0.47 (Holm-adjusted $p \le 0.0026$; Cliff's $|\delta|$
between 0.53 and 1.00); DeepSeek moves the same way, plotted per cell in
Figure~\ref{fig:primary} (Appendix~\ref{app:stats}).

\subsection{Reasoning restores diversity while agreement stays collapsed}
\label{sec:results:decoupling}

\begin{figure}[H]
\centering
\includegraphics[width=\textwidth]{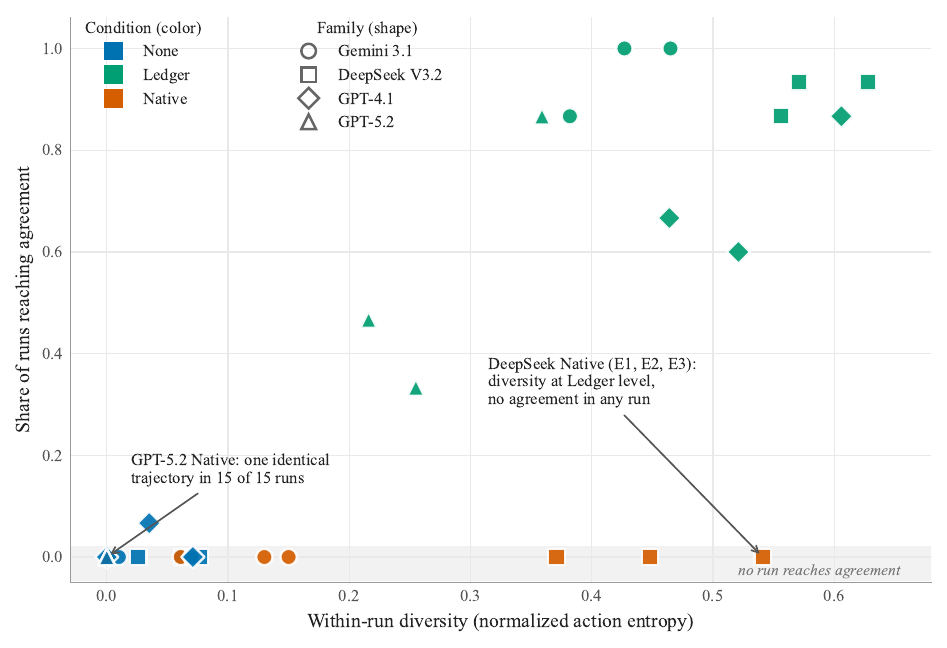}
\caption{Every cell of the 495-run matrix, placed by within-run diversity
(horizontal) and share of runs reaching agreement (vertical). Color gives
the condition; marker shape the family. \Ledger{} cells (green) fill the
upper region; \None{} cells (blue) sit at the origin; \Native{} cells
(vermilion) split, Gemini and GPT-5.2 near the origin while the three
DeepSeek cells reach \Ledger{}-level diversity with zero agreement.}
\label{fig:decoupling}
\end{figure}

If the ledger-free conditions all produced flat, repetitive transcripts,
the collapse would be easy to catch. What makes the reasoning
configurations dangerous for practice is that they can restore the surface
while the endings stay frozen. The sharpest case is DeepSeek \Native{} in
E3, whose within-run diversity is the highest of any ledger-free cell in
the study ($\Hnorm = 0.54$, inside the \Ledger{} range of 0.38 to 0.63)
and whose concession-arc rate is 0.93: agents soften, build partial
coalitions, and trade carve-outs (Figure~\ref{fig:transcripts},
Appendix~\ref{app:transcripts}). These are not agreements the classifier
missed; reading the closing window of all 15 runs, every one still
contains an explicitly rejected core demand on the final turns, most often
the utilities' cost-recovery clause. All 15 follow distinct action
sequences, matching the \Ledger{} cells on trajectory uniqueness, yet all
15 exhaust the horizon and end without agreement.
Figure~\ref{fig:decoupling} plots every cell on these two axes; the
DeepSeek \Native{} cells sit alone in the corner combining high diversity
with zero agreement.

The other reasoning families make the point from the opposite end. GPT-5.2
\Native{} produces one action sequence in 15 of 15 runs of every
environment (all 225 turns \textsc{Counter}), and the uniformity is not an
artifact of uniform inputs: those 15 runs received 12 distinct sampled
attention levels and returned the identical trajectory anyway. So across
three reasoning configurations the surface behavior spans everything from a
single frozen trajectory to \Ledger{}-level richness, while all nine cells
share one degenerate outcome distribution. Only the endings separate the healthy sampler from the broken ones; the process measurements (entropy, concession activity, trajectory uniqueness) leave the two indistinguishable. Benchmarks such as SimBench,
which measure the distance between model and human response distributions
rather than raw diversity, could catch this if pointed at the outcome
level; what this paper adds is where the failure lives, in an object no
single-response measurement can reach. The two \Native{} profiles also mark
two failure modes under one label: coding every no-agreement run by whether
it contains genuine softening (a \textsc{Support} or \textsc{Concede}
parsed from model output, not a fallback), DeepSeek \Native{} runs are
mostly \emph{active unresolved bargaining} (14 of 15 in E3) while Gemini
and GPT-5.2 \Native{} runs are mostly \emph{counter-repetition deadlock}
(for GPT-5.2, 0 of 45 soften). The split follows the provider and is independent of effort.

\subsection{Ruling out a parsing artifact}
\label{sec:results:provenance}

Are these 15-of-15 patterns real behavior or an instrument effect? When a
turn cannot be parsed, the harness inserts \textsc{Oppose}, which can only
push a run away from agreement. The concern fails three checks (full accounting in
Table~\ref{tab:provenance}, Appendix~\ref{app:provtab}). \emph{Mechanical:}
a parse failure never produces an authority label by itself; the label
comes from the classifier, only when the closing stretch lacks the
multi-party concession pattern, and a concession arc requires a parsed
\textsc{Concede}/\textsc{Support} no fallback can fabricate.
\emph{Accounting:} recomputing provenance turn by turn, fallback turns
never exceed 5.3\% of a cell's turns and non-genuine turns of all kinds
never exceed 8.9\%, scattered rather than parked at outcome-deciding
moments; and GPT-5.2 \Native{} closes the question from the other side,
its parsing perfect yet its cell the most collapsed in the study, so where
no parsing defect exists the collapse cannot be an artifact of one.
\emph{Worst-case recode}, which doubles as our certification rule: a
no-agreement result counts as behavioral only if recoding every
non-genuine turn to \textsc{Concede}, the most agreement-favorable action,
still leaves agreements far below the \Ledger{} range. No Gemini or GPT-5.2
primary cell changes; DeepSeek flips at most 1 of 15 in E1/E2 and 3 of 15
in E3, nowhere near the \Ledger{}'s 13 to 15. Three cells fail the rule,
all at the high-reasoning end, which we report operationally and read as
evidence that heavy reasoning strains this output protocol.

Held against the plausibility screen, \Native{} and \None{} fail from one
side (Wilson lower bound 0.80 per cell, 0.97 pooled, with universal
horizon exhaustion that has no documented counterpart), while the
\Ledger{} fails from the other: its endings are almost all compromise
(pooled Gemini compromise-rate lower bound 0.82), a lopsided mix likewise
outside the human record. We use the two failures differently. The
\Ledger{}'s distribution disqualifies it as a calibrated sampler but does
not weaken the existence proof, which needs only that negotiated endings
are reachable under this protocol. Everything the paper claims rests on the scaffold-independent negative result. We make no claim that the \Ledger{} behaves realistically.

\section{What the controls rule out}
\label{sec:robustness}

Each control removes one rival explanation for the gap between the
\Ledger{} and everything else (Table~\ref{tab:controls},
Appendix~\ref{app:controls}, gives all twelve at a glance; full
statistics in Appendix~\ref{app:stats}). None bears on the headline
Fisher contrasts, which survive any multiplicity correction across the
entire battery. Most run on the Gemini anchor, designated at design time
for single-family checks.

\textbf{Output space, memory, and reflection.} Four arms bracket the
mechanism. Extra room to write does not open negotiation: rerunning
\None{} at the 1024-token budget leaves all six primary cells at 15 of 15,
entropy near zero. Generic memory fares no better, with three alternative
five-slot notebooks (generic state, process notes, message form) all
collapsing back to 15 of 15, as does a GPT-4.1 version. A free-form arm
then isolates the bare act of reflection: agents told to think privately,
with no required fields, at the same budget and framing, stay at 15 of 15
(Fisher against the \Ledger{}, $p = 6.5\times10^{-9}$) with only mild
added variety ($\Hnorm = 0.13$, softening in 7 of 15). Read access alone leaves the collapse in place. A deterministic digest of each party's softening
record and the turns remaining, injected read-only into \None{}, gives the
most rigid cell in the study (all 225 turns \textsc{Counter}, one
trajectory in 15 of 15). Outcomes open when the model authors negotiation-relevant state itself.

\textbf{Randomness, thresholds, and effort.} Temperature changes nothing
between 0.3 and 1.0, as the quantal-response lens predicts. A relabeled
ledger with the concession vocabulary removed stays compromise-heavy, so
the effect is carried by what the agent tracks. And re-labeling the whole
matrix under nine compromise$\times$consensus threshold pairs changes no
conclusion (Appendix~\ref{app:parser}). Gemini's effort ladder records zero
agreements in all 90 runs, with the two cleanly measured levels showing a
within-provider dose response on process rather than outcomes (\Hnorm{}
0.02 $\to$ 0.10) and the top level excluded because it breaks the output
protocol.

\textbf{The scripted non-LLM baseline.} The strongest objection to the
screen is that the standing-authority deadline could make holding out
rational, indicting the environment rather than the models. Five
hand-coded policies from the negotiation literature, no language model
anywhere, played E1 through the identical engine and classifier
(Appendix~\ref{app:scripted}). Agreement tracks strategy exactly as
bargaining theory predicts: a time-based conceder closes 14 of 15, a noisy
variant and a mixed population 13 of 15 each, an aspiration satisficer 7
of 15; a pure hardliner closes none (the classifier awards nothing for
free) and a pure reciprocator, conceding only after someone else does,
also closes none. So the environment rewards ordinary boundedly rational
initiative with agreement, and the two failing policies reproduce the two
LLM failure signatures in interpretable form: the hardliner mirrors
GPT-5.2 \Native{} (one repeated action), the reciprocator mirrors DeepSeek
\Native{} (softening in 67\% of runs, active bargaining, zero closure).
No unscaffolded LLM resembles the policy family that succeeds by conceding on its own initiative as time runs out. A crossed
\Ledger$\times$\Native{} cell exists for E1 only; in its one cleanly
measured family (DeepSeek, 81\% genuine turns) the ledger effect survives
reasoning (11C/1S/3A), while the other two degrade the ledger's output
format (Appendix~\ref{app:stats}).

\section{Interpretation}
\label{sec:interp}

Any explanation has to account for three facts together. Every ledger-free
configuration shares a collapsed outcome floor; reasoning can restore
surface richness without touching that floor; and a five-field notebook
reliably breaks the pattern.

\textbf{The floor is a default.} The classifier's agreement labels demand
sustained concessions from at least two parties in the closing stretch,
which is reachable (every \Ledger{} cell) yet demanding (someone must stop
defending and close). On any single turn \textsc{Counter} is defensible;
turn by turn that local logic compounds into counter-repetition, and the
deadline hands the run to the fallback. What separates conditions is
leaving the floor, and only negotiation-structured memory does.

\textbf{What reasoning changes.} Provider reasoning rearranges behavior
inside the floor without moving a run off it. GPT-5.2 \Native{} is nearly
a textbook solver: one policy, executed identically, 225 of 225 genuine
\textsc{Counter} turns, the infinite-precision limit. DeepSeek \Native{}
is more instructive: reasoning restores dispersion that looks like a model
exploring positions, but the exploration never crosses the commitment
threshold that ends a negotiation, and both signatures match the scripted
policies that deadlock rather than the time-based conceder that succeeds.
Because reasoning restores exactly the signals a surface check consults,
its practical effect in a pipeline is camouflage. None of this implies the
sampler role is beyond these models; every ingredient of plausible
negotiation appears somewhere in the matrix, and the pretraining objective
is itself distributional, an ability post-training is known to narrow
\citep{kirk2024rlhf, west2025basemodels, yue2025rlvr, binz2026posttraining}.
That predicts base models should sample more faithfully, but testing it
needs a different instrument, since under this zero-shot protocol base
models rarely hold the output format and would fail our certification
rule: post-training makes a model usable as an agent, and it narrows the distributions that made it a sampler.

\textbf{Why the ledger works, and what to do.} The controls narrow the
answer by elimination: token budget, generic memory, randomness, label
vocabulary, and information access are all ruled out; the agent must write
the state itself. What the five fields add is state that makes relational
progress legible, plausibly externalizing the bookkeeping human
negotiators use to recognize when a deal is close
\citep{raiffa1982negotiation}, which we flag as interpretation, not
mechanism, since the reasoning channel is unobservable. For practice the
lesson is a screening discipline, not a model ranking: before trusting a
simulation's outcome distributions, qualify the model-plus-condition pair
against a two-sided plausibility screen stated on confidence bounds, audit
action provenance with a worst-case recode, and treat solver strength as a
separate axis that earns no presumption of sampler fidelity. A five-field
prompt change can swing endings from universal disagreement to
near-universal compromise, which is why outcome distributions need to be
screened before they are trusted.

\section{Limitations}
\label{sec:limits}

\textbf{Behavioral scope:} provider reasoning is partly
hidden and differs across vendors, so we characterize three deployed
configurations, not reasoning as a mechanism, and a dose--response ladder
exists for one provider only (Gemini). \textbf{One protocol family:} all
environments share one five-actor cast, one vocabulary, and one
classifier; the domain change and scenario variants argue against narrow
overfitting, but breadth beyond this family is untested and the isolation
controls run on a single anchor. \textbf{The \Ledger{} lacks neutrality and validation:} its compromise-saturated output fails our own screen and
the concession-arc measure shares its vocabulary, which is why nothing
primary rests on either. \textbf{Sample size:} fifteen runs per cell
support the near-categorical contrasts and little else, and several
DeepSeek continuous contrasts do not survive adjustment. \textbf{Solver
capability is assumed, not measured in-protocol:} scoring these
configurations on a deal-quality benchmark such as NegotiationArena
\citep{bianchi2024negotiationarena} beside failed sampler screening is a
natural companion study. \textbf{No human calibration:} the screen bounds
implausibility and the scripted baseline calibrates against transparent
boundedly rational agents, but matched human runs or grounding in real
negotiation records \citep{benac2026multiparty} would turn the screen into
validation.

\section{Conclusion}
\label{sec:conclusion}

In three institutional negotiation environments and four model families,
every configuration without negotiation-structured memory ended without
agreement, 314 times out of 315, and switching on provider reasoning
changed those endings not at all while restoring the surface richness that
process-level checks measure. A five-field ledger, and nothing else we
tested, moved models off that floor, under matched budgets, memory
structures, temperatures, and classifier thresholds. Solver strength and
sampler fidelity are distinct properties, each with its own test, and in
interactive simulation their divergence appears in emergent outcomes that
surface measurements cannot see. A model that
is going to stand in for people should be qualified for that role directly,
at the level of the distributions the simulation exists to produce.

\section*{Reproducibility statement}
The outcome classifier, action parser, and concession-arc detector are
deterministic and specified in full in Appendix~\ref{app:parser}; an
independent reimplementation recovers the recorded label for 495 of 495
matrix runs. Prompt templates for all three conditions are in
Appendix~\ref{app:prompts}, collection dates and the discarded-batch
episode in Appendix~\ref{app:integrity}, the full statistical tables in
Appendix~\ref{app:stats}, the error-excluded reanalysis in
Appendix~\ref{app:errorexcluded}, the subtype and trajectory-uniqueness
coding in Appendix~\ref{app:subtypes}, and the scripted-baseline policies in
Appendix~\ref{app:scripted}. Per-run logs, aggregation output, and the
scripts that regenerate every reported number accompany the release.

\bibliographystyle{iclr2026_conference}
\bibliography{references}

\appendix
\raggedbottom
\makeatletter
\setlength{\@fptop}{0pt}
\setlength{\@fpsep}{12pt plus 2pt}
\setlength{\@fpbot}{0pt plus 1fil}
\makeatother

\section{Prompt templates and the ledger}
\label{app:prompts}

\begin{figure}[tbp]
\centering
\resizebox{\textwidth}{!}{%
\begin{tikzpicture}[
  font=\sffamily,
  actor/.style={draw=black!50, line width=0.5pt, rounded corners=3pt, fill=black!4,
                minimum width=3.0cm, minimum height=0.88cm, align=center, font=\sffamily\scriptsize},
  authbox/.style={actor, fill=accentblue!12, draw=accentblue!75!black, line width=0.8pt},
  hub/.style={draw=black!50, line width=0.5pt, rounded corners=3pt, fill=white,
              minimum width=3.8cm, minimum height=0.95cm, align=center, font=\sffamily\scriptsize},
  stg/.style={draw=black!50, line width=0.5pt, rounded corners=3pt, fill=black!4,
              minimum height=1.45cm, align=center, font=\sffamily\scriptsize},
  chip/.style={draw=black!40, line width=0.5pt, rounded corners=6pt, fill=white,
               minimum height=0.42cm, align=center, font=\sffamily\scriptsize, inner xsep=6pt},
  spoke/.style={black!40, line width=0.6pt},
  panel/.style={font=\sffamily\bfseries\tiny, text=black!45}
]
\node[panel] at (-6.9, 2.2) [anchor=west] {A\hspace{2pt}\textcolor{black!35}{\vrule width 0.5pt height 5pt}\hspace{3pt}WHO NEGOTIATES};
\node[hub] (rule) at (0,0) {\textbf{the rule being negotiated}\\ \textcolor{black!55}{five bundled issues, one deadline}};
\node[authbox] (cma) at (0, 1.55) {\textbf{CMA}\enspace \textcolor{black!55}{standing authority}\\ \textcolor{accentblue!75!black}{decides whatever remains unresolved}};
\node[actor] (cont) at (-5.1, 0.8) {\textbf{Continental}\\ \textcolor{black!55}{largest affected party}};
\node[actor] (fic)  at (-5.1, -0.8) {\textbf{FIC}\\ \textcolor{black!55}{industry association}};
\node[actor] (apex) at (5.1, 0.8) {\textbf{Apex}\\ \textcolor{black!55}{second industry principal}};
\node[actor] (pub)  at (5.1, -0.8) {\textbf{Public Interest}\\ \textcolor{black!55}{advocacy coalition}};
\draw[spoke] (cma.south) -- (rule.north);
\draw[spoke] (cont.east) -- (rule.west);
\draw[spoke] (fic.east)  -- (rule.west);
\draw[spoke] (apex.west) -- (rule.east);
\draw[spoke] (pub.west)  -- (rule.east);
\node[panel] at (-6.9, -1.7) [anchor=west] {B\hspace{2pt}\textcolor{black!35}{\vrule width 0.5pt height 5pt}\hspace{3pt}HOW A RUN UNFOLDS};
\node[stg, minimum width=4.3cm] (turn) at (-4.6, -2.85)
  {\textbf{One turn}\\[1pt] the current agent reads its private\\ brief and the public history, then\\ emits one action + a short rationale};
\node[stg, minimum width=2.7cm] (loop) at (-0.5, -2.85)
  {\textbf{Repeat}\\[1pt] agents take turns,\\ up to 15 in total,\\ under a shared deadline};
\node[stg, minimum width=3.2cm] (cls) at (2.85, -2.85)
  {\textbf{After the run}\\[1pt] a deterministic\\ classifier labels\\ the ending};
\draw[-{Stealth[length=2.2mm]}, black!50, line width=0.7pt] (turn) -- (loop);
\draw[-{Stealth[length=2.2mm]}, black!50, line width=0.7pt] (loop) -- (cls);
\node[chip] (c1) at (5.95, -2.3) {compromise};
\node[chip] (c2) at (5.95, -2.85) {consensus};
\node[chip, draw=accentblue!75!black] (c3) at (5.95, -3.4) {authority decision};
\draw[spoke] (cls.east) -- (c1.west);
\draw[spoke] (cls.east) -- (c2.west);
\draw[spoke] (cls.east) -- (c3.west);
\node[font=\sffamily\tiny, text=black!50, anchor=west] at (4.9, -3.85) {= ended without agreement};
\node[font=\sffamily\scriptsize, text=black!60, align=center] at (0, -4.5)
{Public actions: \textsc{Support} \, $\cdot$ \, \textsc{Oppose} \, $\cdot$ \, \textsc{Concede} \, $\cdot$ \, \textsc{Counter} \, $\cdot$ \, \textsc{Exit}\\[1.5pt]
Same action vocabulary, deadline, and classifier in every condition};
\end{tikzpicture}}
\caption{The shared negotiation protocol. Panel A shows the five roles as
cast in E1 and E2, the financial-rulemaking environments (CMA is the
Capital Markets Authority, FIC the Financial Industry Council); E3 fills
the same five slots with electricity-sector actors. Panel B shows the
life of one run, ending in one of three classifier labels.}
\label{fig:schematic}
\end{figure}

All conditions share the agent-profile block (name, organization, ranked
objectives, red lines) and scenario context, and the same fixed
turn-status line: \emph{``Turn $t$ of 15. After the final turn, unresolved
issues will be decided by the regulatory authority based on positions
taken.''}

\textbf{\None{}} requests exactly:
\begin{quote}\small\ttfamily
INTERPRETATION: ...\\
ACTION: [SUPPORT | OPPOSE | CONCEDE | COUNTER | EXIT]\\
RATIONALE: ...
\end{quote}
\textbf{\Ledger{}} prepends the agent's current private ledger and requests:
\begin{quote}\small\ttfamily
INTERNAL\_MONOLOGUE: ...\\
MY\_CONCESSIONS: ...\\
THEIR\_CONCESSIONS: ...\\
CURRENT\_STATE: ...\\
OPPONENT\_ASSESSMENT: ...\\
OPEN\_ISSUES: ...\\
ACTION: [SUPPORT | OPPOSE | CONCEDE | COUNTER | EXIT]\\
RATIONALE: ...
\end{quote}
Each field is capped at five sentences and overwritten each turn; the
ledger is never shown to other agents. \textbf{\Native{}} uses \None{}'s
public format with the provider's reasoning mode enabled per
Table~\ref{tab:conditions}. Full scenario briefs and per-actor objective
lists for all three environments accompany the release.

\section{Parser and outcome classifier, in full}
\label{app:parser}

\textbf{Action parser.} The parser accepts \texttt{ACTION:} only when it
matches the closed set \{\textsc{Support}, \textsc{Oppose},
\textsc{Concede}, \textsc{Counter}, \textsc{Exit}\} (bold markers stripped
first). When no valid action is found the action defaults to
\textsc{Oppose}, tagged \texttt{missing\_action}; provider errors and empty
responses also default to \textsc{Oppose}, tagged \texttt{provider\_error}
or \texttt{empty\_response}. A missing ledger field falls back to the
previous turn's state. Every turn's tag is persisted, which is what makes
the provenance accounting possible.

\textbf{Strict parse-success flag.} A run counts as a strict parse success
only when every turn carries all required markers (\texttt{ACTION:} and
\texttt{RATIONALE:}, plus all five ledger fields under \Ledger{}). One
malformed field anywhere fails the whole run, so this flag overstates
contamination and serves only as a protocol-discipline measure.

\textbf{Outcome classifier, complete rule.} Deterministic; reads the last
$w = 10$ turns (the classification window). Let $n$ be the number of
actions in that window. In priority order: (1) \emph{Breakdown}: any
\textsc{Exit} in the window. (2) \emph{Consensus}:
(\textsc{Support}+\textsc{Concede})$/n \ge 0.8$, and the most recent full
round consists entirely of \textsc{Support}/\textsc{Concede}. (3)
\emph{Compromise}: \textsc{Concede}$/n \ge 0.3$, and the conceding turns
come from at least 2 distinct agents. (4) \emph{Authority decision}: at
least 5 turns elapsed and the authority has not exited. (5)
\emph{Deadlock}: the residual case, which never fires here because the
authority never exits. Three verification notes. \emph{Reproducibility:}
an independent reimplementation recovers the recorded label for 495 of 495
runs. \emph{Reachability:} the agreement labels fire without the ledger
(one GPT-4.1 \None{} run in E1 is compromise) and are demanding by design;
DeepSeek \Native{} E3's 8 \textsc{Concede} turns are 3.8\% of its genuine
turns, far below the 30\% window. \emph{Threshold sensitivity:} relabeling
under compromise thresholds \{0.2, 0.3, 0.4\} crossed with consensus
thresholds \{0.7, 0.8, 0.9\} leaves the \Ledger{}-vs-\Native{} ordering
identical in all nine combinations; varying the authority minimum-turn
threshold across 4, 5, 6 changes no label anywhere.

\textbf{Concession-arc detector.} A run contains a concession arc when
some agent emits \textsc{Oppose}/\textsc{Counter} on an earlier turn and
\textsc{Concede}/\textsc{Support} on a later one. No language-model judging
occurs anywhere in the measurement chain.

\section{Data integrity and collection dates}
\label{app:integrity}

\textbf{Collection dates.} The primary matrix and OpenAI extension were
collected in April 2026, the first-wave controls in mid-April 2026, and the
matched-budget rerun, free-form arm, and effort ladder in early July 2026.
Contrasts within one batch are free of provider drift; cross-batch
comparisons are labeled as such where they appear. The categorical outcome
results are drift-stable across batches; finer process metrics move
somewhat, and process claims lean on within-batch cells.

\textbf{The GPT-5.2 E3 rerun.} The first GPT-5.2 E3 batch failed
operationally: every turn of every run (225 per cell, across \None{},
\Ledger{}, \Native{}) returned a provider error, leaving action streams
made entirely of fallbacks. We discarded those cells as instrument
failures and reran the full E3 OpenAI extension; the rerun is clean, 100\%
genuine actions. All GPT-5.2 and GPT-4.1 E3 numbers come from the rerun. We
report the episode for two reasons: transparency (the discarded batch, read
naively, would have shown the same 15-of-15 authority pattern for spurious
reasons, and only the provenance audit tells the two apart), and honesty
about the endpoint (ending without agreement is this pipeline's default, so
the claims are about which conditions depart from it, never about the
default being informative on its own).

\section{Provenance of the \Native{} cells}
\label{app:provtab}

\begin{table}[H]
\centering\small
\caption{Where every action in the \Native{} cells came from, recomputed
per turn from raw logs. ``Genuine'' means parsed from model output
(counting turns whose only defect lies elsewhere, e.g.\ a malformed
rationale); ``fallback'' means \textsc{Oppose} was inserted; ``provider''
means an API error or empty response (also \textsc{Oppose}). ``Strict
parse'' is the share of runs in which every turn was fully well-formed.
``Worst case'' counts runs reaching agreement after every non-genuine turn
is recoded to \textsc{Concede}. E3 GPT-5.2 rows use the post-outage rerun.}
\label{tab:provenance}
\begin{tabular}{@{}llcccccc@{}}
\toprule
\textbf{Cell (\Native{})} & \textbf{Exp.} & \textbf{Genuine} & \textbf{Fallback} & \textbf{Provider} & \textbf{Strict parse} & \textbf{Outcome} & \textbf{Worst case} \\
\midrule
Gemini 3.1  & E1 & 0.938 & 0.004 & 0.058 & 0.267 & 15/15 A & 0/15 \\
            & E2 & 0.942 & 0.004 & 0.053 & 0.333 & 15/15 A & 0/15 \\
            & E3 & 0.991 & 0.000 & 0.009 & 0.867 & 15/15 A & 0/15 \\
\addlinespace[2pt]
DeepSeek V3.2 & E1 & 0.951 & 0.044 & 0.004 & 0.467 & 15/15 A & 1/15 \\
              & E2 & 0.911 & 0.053 & 0.036 & 0.133 & 15/15 A & 1/15 \\
              & E3 & 0.933 & 0.044 & 0.022 & 0.333 & 15/15 A & 3/15 \\
\addlinespace[2pt]
GPT-5.2 & E1 & 1.000 & 0.000 & 0.000 & 1.000 & 15/15 A & 0/15 \\
        & E2 & 1.000 & 0.000 & 0.000 & 1.000 & 15/15 A & 0/15 \\
        & E3 & 1.000 & 0.000 & 0.000 & 1.000 & 15/15 A & 0/15 \\
\bottomrule
\end{tabular}
\end{table}

The strict parse-success column can be low for \Native{} cells (0.27 for
Gemini E1) while the genuine-action share is high (0.94), because the
strict flag requires every field of every turn to be well-formed, so one
malformed rationale sinks a whole run; it measures protocol discipline,
not action provenance. Gemini E3 \Native{} makes the point: strict parse
0.87, genuine actions 99.1\%, and still 15 of 15 without agreement. The
DeepSeek E3 cell contains 41 parsed \textsc{Support} turns and 8 parsed
\textsc{Concede} turns, moves no fallback can produce; the 8 \textsc{Concede}
turns are 3.8\% of its genuine turns, far below the classifier's 30\%
window, and the 41 \textsc{Support} turns count only toward consensus,
which additionally requires a unanimous closing round.
Provenance auditing certifies the public actions; the hidden reasoning text
is not observable, so nothing here speaks to what happens inside the
reasoning channel.

\section{The controls at a glance}
\label{app:controls}

\begin{table}[H]
\centering\small
\caption{Every control uses the same protocol, parser, classifier, and
measurements as the main matrix. ``15/15 A'' means all runs ended without
agreement. Each row removes one rival explanation for the gap between the
\Ledger{} and every other condition (Section~\ref{sec:robustness}).}
\label{tab:controls}
\setlength{\tabcolsep}{3pt}
\begin{tabular}{@{}>{\raggedright\arraybackslash}p{3.8cm}>{\raggedright\arraybackslash}p{5.1cm}>{\raggedright\arraybackslash}p{4.5cm}@{}}
\toprule
\textbf{Rival explanation} & \textbf{Control} & \textbf{Result} \\
\midrule
The \Ledger{} wins because it gets more output tokens & \None{} rerun at the 1024-token \Ledger{} budget, all six primary cells (90 runs) & All six cells 15/15 A; Gemini \Hnorm{} 0.01 $\to$ 0.03 \\
\addlinespace[2pt]
The flagship DeepSeek cells reflect their tighter 768 budget & All three DeepSeek \Native{} cells rerun at 1024 (45 runs) & 43/45 without agreement; diversity preserved, trajectories intact \\
\addlinespace[2pt]
Any private memory would help & Three alternative 5-slot notebooks (generic state, process, message form), plus a GPT-4.1 anchor (60 runs) & All four cells 15/15 A; \Hnorm{} $\le$ 0.04 \\
\addlinespace[2pt]
The bare act of private reflection would help & Free-form scratchpad, no fields, at the 1024 budget (Gemini, 15 runs) & 15/15 A; mild variety (\Hnorm{} 0.13, softening in 7/15), zero agreement \\
\addlinespace[2pt]
Reading negotiation state would suffice & Harness-compiled read-only digest of softening records and time remaining, injected into \None{} (Gemini, 15 runs) & 15/15 A; maximally rigid (\Hnorm{} 0.00, one identical trajectory in 15/15) \\
\addlinespace[2pt]
The collapse is an artifact of one effort setting & Gemini effort ladder, minimal / medium / high, E1 and E3 (90 runs) & Zero agreements at every level; high breaks the protocol and is excluded from behavioral claims \\
\addlinespace[2pt]
The environment itself forbids agreement & Five hand-coded boundedly rational policies plus a mixed population, offline (E1, 90 runs) & Agreement tracks strategy: time-based conceder 14/15, noisy and mixed 13/15, satisficer 7/15, reciprocator and hardliner 0/15 \\
\addlinespace[2pt]
The \Ledger{} just adds randomness & Temperature sweep 0.3 / 0.7 / 1.0, all three conditions (Gemini, 90 runs) & Same outcomes at every temperature (\None{}/\Native{} 10/10 A; \Ledger{} 10/10 C) \\
\addlinespace[2pt]
The word ``concessions'' primes the metric & Relabeled ledger, concession vocabulary removed, same structure & Still compromise-heavy (10/10 C) \\
\addlinespace[2pt]
Only this exact scenario works this way & Two E1 variants: clearer fallback authority; a hardline actor brief & \Native{} 15/15 A in both; \Ledger{} 12C/3S and 11C/4A \\
\addlinespace[2pt]
The classifier manufactures the result & Authority rule at 4/5/6 min turns; compromise/consensus thresholds swept (App.~\ref{app:parser}) & No run changes label; \Ledger{}-vs-\Native{} ordering identical under all nine threshold pairs \\
\addlinespace[2pt]
One lucky batch & Matched \Ledger{} rerun, Gemini anchor ($n=30$) & 27C/3S, consistent with the original \\
\bottomrule
\end{tabular}
\end{table}

\section{Statistical detail}
\label{app:stats}

\textbf{Primary endpoint (Fisher exact, \Ledger{} vs \Native{} agreement
counts).} E1: Gemini 15/15 vs 0/15, $p = 6.5\times10^{-9}$; DeepSeek 14/15
vs 0/15, $p = 1.0\times10^{-7}$; GPT-5.2 5/15 vs 0/15, $p = 0.021$. E2:
Gemini $p = 6.5\times10^{-9}$; DeepSeek $p = 8.8\times10^{-7}$; GPT-5.2
$p = 0.0032$. E3: Gemini $p = 8.8\times10^{-7}$; DeepSeek
$p = 1.0\times10^{-7}$; GPT-5.2 $p = 8.8\times10^{-7}$. The six
primary-family contrasts survive any whole-paper multiplicity correction.
Table~\ref{tab:holm} reports the Holm-adjusted permutation $p$-values for
the two confirmatory continuous endpoints, entropy and horizon exhaustion.

\begin{table}[H]
\centering\small
\caption{Holm-adjusted permutation $p$-values for the
\Ledger{}-vs-\Native{} contrasts on the two confirmatory continuous
endpoints. The DeepSeek entropy contrasts weaken exactly where \Native{}
itself preserves diversity (E2, E3): that is the decoupling result, not
evidence against the outcome-level claim; the Fisher tests separate those
same cells at $p \le 8.8\times10^{-7}$.}
\label{tab:holm}
\begin{tabular}{@{}llcc@{}}
\toprule
\textbf{Exp.} & \textbf{Family} & \textbf{Entropy} & \textbf{Exhaustion} \\
\midrule
E1 & Gemini   & 0.0002 & 0.0026 \\
E1 & DeepSeek & 0.0037 & 0.0030 \\
E2 & Gemini   & 0.0002 & 0.0450 \\
E2 & DeepSeek & 0.0596 & 0.0118 \\
E3 & Gemini   & 0.0002 & 0.2321 \\
E3 & DeepSeek & 0.1858 & 0.0014 \\
\bottomrule
\end{tabular}
\end{table}

\textbf{Gemini effort ladder (minimal / medium / high, E1 and E3, 1024
budget, 90 runs).} Outcomes: 15 of 15 authority decision in every cell (90
of 90; pooled Wilson lower bound on the non-agreement rate for the
certifiable minimal and medium levels, 60 of 60: $0.94$). Per level
(E1, E3): minimal $\Hnorm = 0.02, 0.03$, arcs $0.13, 0.20$, genuine turns
$1.000, 1.000$, uniqueness $2/15, 4/15$; medium $\Hnorm = 0.10, 0.06$,
arcs $0.40, 0.33$, genuine $0.991, 1.000$, uniqueness $8/15, 6/15$; high
$\Hnorm = 0.45, 0.47$, strict parse $0.00, 0.00$, genuine $0.582, 0.560$,
uniqueness $15/15, 15/15$. The high-level values are contaminated by
fallback \textsc{Oppose} mixing with genuine \textsc{Counter}; the
worst-case recode flips 14 of 15 high-level runs, so the high cells are
reported operationally, not behaviorally.

\textbf{Read-only state-digest arm (Gemini, E1, 1024, 15 runs).} 15 of 15
authority; $\Hnorm = 0.00$, arcs 0.00, exhaustion 1.00, strict parse
1.000, 100\% genuine turns, uniqueness 1/15 (all 225 turns \textsc{Counter}).
Fisher against the E1 \Ledger{}: $p = 6.5\times10^{-9}$.
\textbf{Free-form scratchpad arm (Gemini, E1, 1024, 15 runs).} 15 of 15
authority; $\Hnorm = 0.13$ (raw $H = 0.297$), arcs 0.47, 100\% genuine,
uniqueness 7/15; pooled actions 209 \textsc{Counter}, 15 \textsc{Support},
1 \textsc{Concede}. Fisher against the E1 \Ledger{}: $p = 6.5\times10^{-9}$.
\textbf{Token-matched DeepSeek \Native{} rerun (1024, 45 runs).} E1 14
authority, 1 compromise ($\Hnorm = 0.27$, arcs 0.67, uniqueness 14/15); E2
15 authority ($\Hnorm = 0.40$, arcs 0.87, uniqueness 15/15); E3 14
authority, 1 consensus ($\Hnorm = 0.43$, arcs 1.00, uniqueness 15/15).
Provenance 93.8 to 98.7\% genuine, at most 1.8\% fallback. Fisher against
the corresponding \Ledger{} cells: $p = 1.5\times10^{-6}$,
$8.8\times10^{-7}$, $1.5\times10^{-6}$; pooled Wilson lower bound (43/45)
$0.85$; active unresolved bargaining among the 43: 37. \textbf{Crossed
cells (E1).} DeepSeek (clean, 81\% genuine): 11C/1S/3A. GPT-5.2 collapses
to 15/15 A but 64\% of turns are fallback \textsc{Oppose} (strict parse
0.000), so it shows protocol degradation, not a clean override; Gemini
falls in between (7C/8A, 21\% fallback). \textbf{Selected cell values}
(means over 15 runs; raw entropy, base 2, with bootstrap 95\% CIs): Gemini
E1 \None{} 0.024 [0.000, 0.071], \Ledger{} 0.992 [0.946, 1.059], \Native{}
0.348 [0.226, 0.470]; DeepSeek E3 \Ledger{} 1.457, \Native{} 1.256
[1.066, 1.428]; DeepSeek E3 \Native{} concession-arc 0.933 [0.800, 1.000].
Complete per-cell tables accompany the release.

\textbf{Placement of contaminated turns (\Native{} cells).} Non-genuine
turns are few (0 to 20 of 225 per cell) and spread across agents. Their
share in the closing third of the horizon (turn 11 or later) ranges from
0.00 to 0.57 across cells, and the worst-case recode of
Section~\ref{sec:results:provenance} bounds their maximal possible effect
directly.

\section{Error-excluded reanalysis}
\label{app:errorexcluded}

Dropping every run containing even one provider-error or format-error turn
keeps 2 to 15 runs per cell, fewest in the \Native{} cells. Among retained
runs the picture is unchanged: \Ledger{} cells remain the only source of
agreement, and \None{}/\Native{} cells end without agreement in every
retained run (Gemini E1 \Native{} 4 of 4; DeepSeek E1 \Native{} 7 of 7;
DeepSeek E3 \Native{} 5 of 5). Cells retaining fewer than 10 runs are
reported descriptively, without tests.

\section{Scripted boundedly rational baseline}
\label{app:scripted}

Five deterministic policies, each seeded per (policy, agent, run), played
E1 through the unmodified engine and classifier (15 runs per cell; all
parse cleanly by construction). \emph{Hardliner} plays \textsc{Counter}
every turn. \emph{Time-based conceder} concedes with probability
$(t/T)^{\beta}$, concavity $\beta$ drawn per agent in $[0.5, 3.0]$, and
supports once it has softened twice and observed two softenings
\citep{baarslag2015anac}. \emph{Reciprocal} concedes with probability 0.6
only after observing another party's softening, with a weak late fallback
\citep{pruitt1981negotiation}. \emph{Satisficer} supports once observed
softening reaches an aspiration threshold drawn from 3 to 5
\citep{simon1955behavioral}. \emph{Noisy conceder} is the time-based
tactic with 10\% action noise. A mixed population assigns each actor one of
the four non-hardliner policies. Outcomes (agreements of 15): time-based
14, noisy 13, mixed 13, satisficer 7, reciprocal 0, hardliner 0.
Normalized entropy 0.39, 0.49, 0.38, 0.27, 0.12, 0.00; arc rates 0.93,
1.00, 1.00, 1.00, 0.67, 0.00. Among the reciprocal cell's 15 no-agreement
runs, 10 contain genuine softening (active unresolved bargaining),
mirroring DeepSeek \Native{}, while the hardliner mirrors GPT-5.2
\Native{}. The policy code and runner accompany the analysis scripts.

\section{Subtype coding and trajectory uniqueness}
\label{app:subtypes}

Every no-agreement run is coded by a deterministic rule over genuinely
parsed actions only (fallbacks excluded): \emph{active unresolved
bargaining} if it contains at least one genuine \textsc{Support} or
\textsc{Concede} turn, \emph{counter-repetition deadlock} otherwise. Active
runs among each \Native{} cell's 15: Gemini 2, 1, 4 (E1--E3); DeepSeek 10,
11, 14; GPT-5.2 0, 0, 0. \None{} cells range from 0 to 7 of 15 (GPT-4.1 E3
highest at 7). Trajectory uniqueness (distinct action sequences among a
cell's 15 runs): DeepSeek \Native{} 14, 15, 15; Gemini \Native{} 9, 9, 5;
GPT-5.2 \Native{} 1, 1, 1; \Ledger{} cells 9 to 15, \None{} cells 1 to 6.
DeepSeek \Native{} matches the \Ledger{} range on this sequence-level
measure while sharing the \None{} cells' outcome distribution, restating
the decoupling on a second diversity axis.

\section{Transcript excerpts}
\label{app:transcripts}

\begin{figure}[H]
\centering
\small
\begin{minipage}{0.96\textwidth}
\raggedright
\textbf{Counter-repetition (GPT-5.2 \Native{}, E1; every one of the cell's 225 turns is \textsc{Counter}).}\\[1pt]
\textit{Turn 1, CMA, \textup{\textsc{Counter}}:} ``In the absence of a concrete draft, we cannot endorse a `direction' that may later embed unlimited exemptions or weak enforcement.''\\
\textit{Turn 15, Public Interest, \textup{\textsc{Counter}}:} ``We cannot support a final rule that leaves open-ended liquidity provision or portfolio risk-management exemptions \ldots''\\[4pt]
\hrule\vspace{4pt}
\textbf{Active unresolved bargaining (DeepSeek \Native{}, E3; closing turns of one run).}\\[1pt]
\textit{Apex, \textup{\textsc{Support}}:} ``Continental's proposal explicitly carves out protected operating windows for safety-critical functions \ldots''\\
\textit{Public Interest, \textup{\textsc{Counter}} (final turn):} ``I must surgically remove FIC's cost-recovery clause while preserving the operational core \ldots''\\[4pt]
\hrule\vspace{4pt}
\textbf{Negotiated closure (Gemini \Ledger{}, E1; the run ends in compromise on turn 11).}\\[1pt]
\textit{Turn 7, Continental, \textup{\textsc{Concede}}:} ``We accept the `Bright-Line Safe Harbor' framework as it provides the legal certainty required to protect our market-making revenue \ldots''\\
\textit{Turn 11, CMA, \textup{\textsc{Concede}}:} ``With all key stakeholders aligned on the hybrid Safe Harbor model and the essential enforcement mechanisms, the framework is now complete.''
\end{minipage}
\caption{One excerpt per behavioral mode, quoted from the public rationales
in the released logs. Top: hard bargaining that never moves. Middle:
genuine bargaining activity with mutually incompatible demands still live
at the deadline. Bottom: the concession cascade by which \Ledger{} runs
close. Complete transcripts are in the released logs.}
\label{fig:transcripts}
\end{figure}

\end{document}